\documentclass[10pt,twocolumn,letterpaper]{article}

\usepackage[pagenumbers]{cvpr} 

\usepackage{graphicx}
\usepackage{amsmath}
\usepackage{amssymb}
\usepackage{booktabs}
\usepackage{multirow}
\usepackage{float}
\usepackage{xcolor,colortbl}
\definecolor{Gray}{gray}{0.9}
\definecolor{LightCyan}{rgb}{0.9,1,1}
\usepackage{mathtools}
\DeclarePairedDelimiter{\norm}{\lVert}{\rVert} 

\usepackage[pagebackref,breaklinks,colorlinks]{hyperref}

\usepackage[capitalize]{cleveref}
\crefname{section}{Sec.}{Secs.}
\Crefname{section}{Section}{Sections}
\Crefname{table}{Table}{Tables}
\crefname{table}{Tab.}{Tabs.}

\def\cvprPaperID{3560} 
\def\confName{CVPR}
\def\confYear{2022}

\begin{document}

\title{Rethinking Query, Key, and Value Embedding in Vision Transformer \\under Tiny Model Constraints}

\author{Jaesin Ahn$^\dag$, Jiuk Hong$^\dag$, Jeongwoo Ju$^\ddag$ and Heechul Jung$^\dag$\\
$^\dag$Department of Artificial Intelligence, Kyungpook National University\\
$^\ddag$Korea Advanced Institute of Science and Technology\\
{\tt\small \{ajs0420, hong4497, heechul\}@knu.ac.kr, veryju@kaist.ac.kr}
}
\maketitle

\begin{abstract}
A vision transformer (ViT) is the dominant model in the computer vision field.
Despite numerous studies that mainly focus on dealing with inductive bias and complexity, there remains the problem of finding better transformer networks.
For example, conventional transformer-based models usually use a projection layer for each query (Q), key (K), and value (V) embedding before multi-head self-attention.
Insufficient consideration of semantic $Q, K$, and $V$ embedding may lead to a performance drop.
In this paper, we propose three types of structures for $Q$, $K$, and $V$ embedding. The first structure utilizes two layers with ReLU, which is a non-linear embedding for $Q, K$, and $V$. The second involves sharing one of the non-linear layers to share knowledge among $Q, K$, and $V$. The third proposed structure shares all non-linear layers with code parameters. The codes are trainable, and the values determine the embedding process to be performed among $Q$, $K$, and $V$.
Hence, we demonstrate the superior image classification performance of the proposed approaches in experiments compared to several state-of-the-art approaches.
The proposed method achieved $71.4\%$ with a few parameters (of $3.1M$) on the ImageNet-1k dataset compared to that required by the original transformer model of XCiT-N12 ($69.9\%$). Additionally, the method achieved $93.3\%$ with only $2.9M$ parameters in transfer learning on average for the CIFAR-10, CIFAR-100, Stanford Cars datasets, and STL-10 datasets, which is better than the accuracy of $92.2\%$ obtained via the original XCiT-N12 model.
\end{abstract}

\section{Introduction}
\label{sec:intro}

\begin{figure}[ht]
    \centering\includegraphics[width=\linewidth]{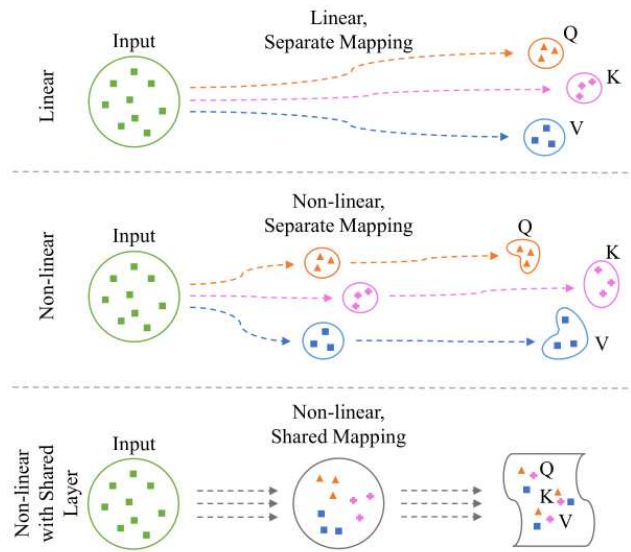}
    \caption{Conceptual diagram of conventional linear mapping, our non-linear mapping, and our non-linear mapping with shared layer. Compared to separate mapping, shared mapping can find a new combination of $Q, K$, and $V$ in a shared manifold.}
    \vspace{-5mm}
    \label{fig_mapping}
\end{figure}

Currently, transformers have emerged as dominant models in deep learning because of their superior performance, especially in terms of long-range dependency~\cite{Vaswani2017-yi, Devlin2018-fl}.
Transformers first appeared in natural language processing (NLP)~\cite{Vaswani2017-yi} and have since become a widely used backbone network of state-of-the-art models~\cite{Radford2018-oc, Radford2019-gr, brown2020language}.
A transformer mainly adopts a self-attention mechanism with the following feedforward network to capture the sequence of input tokens.
The self-attention mechanism can deal with the global information of input tokens, and this feature allows the models to become state-of-the-art models~\cite{brown2020language, Lewis2019-wd}.

Although the use of transformers have been increasing in NLP, convolutional neural networks (CNN) have been remained in the mainstream in computer vision (CV)~\cite{Szegedy2015-sb, He2016-wt, Zoph2018-yd, Tan2019-zx}.
CNN has representative features such as translation equivariance and spatial locality.
These features enable the dominant performance of CNNs in CV tasks~\cite{Tan2019-zx, ronneberger2015u, Redmon2016-za, Howard2019-tj} such as image classification, object detection, and segmentation; however, recently, transformers have also increased in CV because vision transformers (ViT)~\cite{dosovitskiy2020image} offer competitive performance.
As current research shows ~\cite{Carion2020-xo, dosovitskiy2020image, Touvron2020-io}, transformer-based networks have now become state-of-the-art models replacing CNN-based networks.

ViT shows superior performance by removing inductive bias, which is a representative feature of CNN and reinforcing long-range dependency.
Standard vision transformers flatten 2-dimensional images into 1-dimensional sequences of tokens and then apply a global self-attention mechanism to extract the relation among tokens.
This procedure can consider the long-range relationship among tokens, in contrast to CNNs, which consider the local relationship of image pixels.
However, it results in extensive computations that increase quadratically according to the image resolution and the need for pre-trained weights~\cite{dosovitskiy2020image, steiner2021train} using large-scale datasets, that is, JFT-300M~\cite{sun2017revisiting}, ImageNet-21k~\cite{Deng2009-eo}.
To alleviate these inefficiencies, several researchers have considered adding the inductive bias in the transformer explicitly~\cite{wu2021cvt, yuan2021tokens, yang2021focal, yuan2021volo}, and attempted to reduce the computational cost by modifying the multi-head self-attention (MHSA) method~\cite{El-Nouby2021-ms}.
However, there are few considerations regarding new embedding techniques for query ($Q$), key ($K$), and value ($V$) \cite{wu2021cvt}.

In this study, we consider the design of the structure of embedding $Q$, $K$, and $V$. First, we present a two-layer embedding structure model with a rectified linear unit activation function (ReLU). In contrast to the original embedding, the structure can transform the input data into a non-linear space. It has the potential to improve the performance of the ViT because more non-linearity can solve more complicated problems. The second structure is a one-layer shared structure, which is a variant of the first structure. The third structure is a two-layer shared structure with code parameters to improve the original self-attention mechanism used in conventional transformers.
Finally, experimental results demonstrate that the proposed method outperforms the state-of-the-art (SOTA) model in the ImageNet~\cite{Deng2009-eo} classification task and transfer learning on multiple datasets (i.e., CIFAR-10, CIFAR-100~\cite{krizhevsky2009learning}, Stanford Cars~\cite{Krause2013-ui}, and STL-10~\cite{coates2011analysis}) for evaluating various image classification tasks.

\section{Related Works}
ViT was first proposed in Dosovitskiy \etal.~\cite{dosovitskiy2020image}.
This pioneering work directly applies a plain transformer encoder, which is commonly used in NLP.
They split images into patches to embed them as tokens and feed them to the transformer encoder with additional positional embeddings.
The transformer encoder consists of repeated multi-head self-attention and multilayer perceptron (MLP) layers.
In contrast to CNN structures, the transformer can handle long-range information by applying global self-attention among the tokens.
However, global self-attention causes a lack of inductive bias, which makes large-scale pre-training necessary and quadratically increasing computation.
Although DeiT~\cite{Touvron2020-io} suggests distilling knowledge using the CNN model as a teacher model to train the transformer without inductive bias, it requires an additional teacher model that causes extra computation.

Therefore, current research mainly focuses on applying explicit inductive bias to vision transformers ~\cite{xu2021vitae, d2021convit, li2021localvit, yuan2021tokens} and reducing the computational complexity from quadratic to linear~\cite{jaegle2021perceiver, El-Nouby2021-ms}.
XCiT~\cite{El-Nouby2021-ms} is one study that focused on reducing the computational complexity.
It introduces cross-covariance attention (XCA) instead of standard self-attention.
XCA effectively reduces computation without a large performance gap compared to the baseline.
However, it does not consider $Q, K$, and $V$ embedding, which directly affects the attention operation.
In contrast to these works, we introduce $Q, K$, and $V$ embedding techniques to improve the performance of image recognition.

CvT~\cite{wu2021cvt} is the most similar work conducted before.
CvT utilizes convolutional projection for the $Q, K$, and $V$ embedding of the transformer encoder to take advantage of both the CNN and transformer.
However, they also stay to use separate projections and linear operations.
This may lead to a constrained combination of $Q, K$, and $V$.
Although CvT-13 achieved 81.6\% in ImageNet classification using 20M parameters and larger models achieved higher scores, they did not consider tiny model constraints.

\section{Attention Mechanism in Vision Transformer}
ViT~\cite{dosovitskiy2020image} constructs the network following the original NLP transformer encoder.
It splits images into sets of patches and uses them as tokens, which are embedded with the positional embedding.
The embedded vector is used for the following repeated self-attention and feed-forward networks. In this section, we introduce two representative approaches to the self-attention mechanism, namely self-attention and cross-covariance attention.

\begin{figure*}[!ht]
    \centerline{
        \subfloat[Conventional]{
            \includegraphics[width=0.24\textwidth]{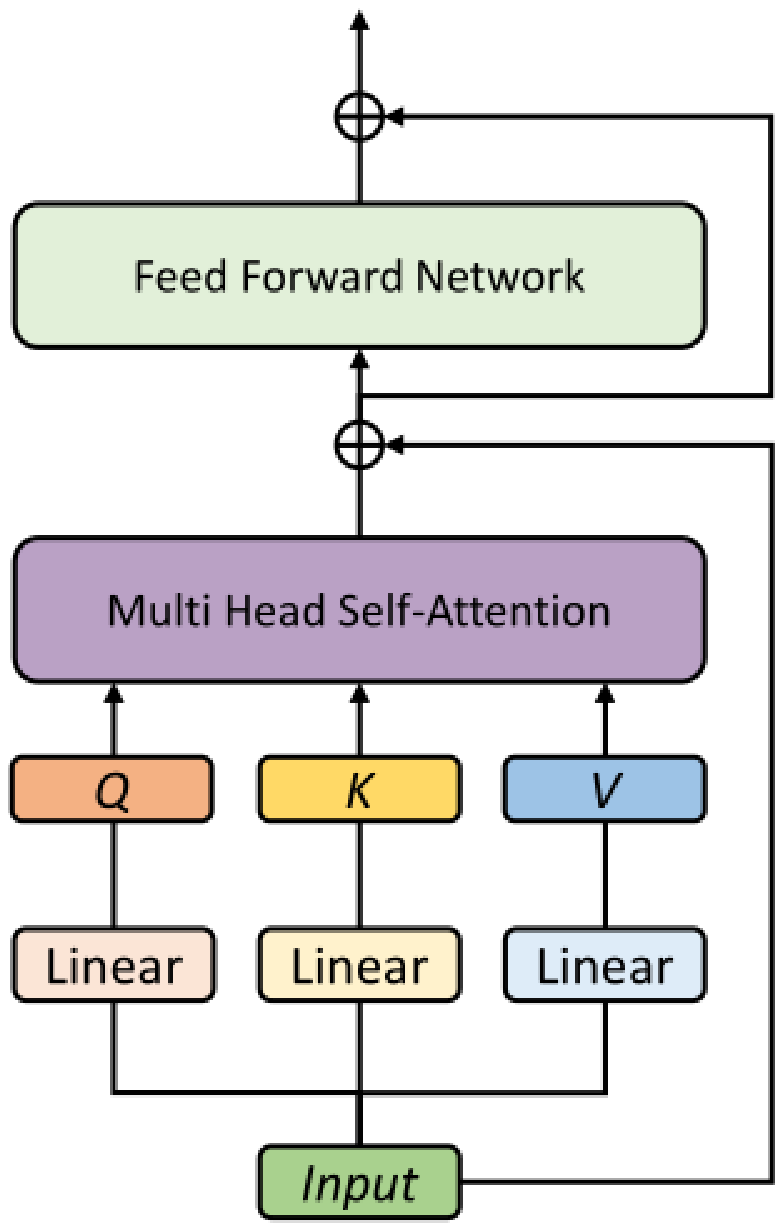}%
            \label{fig_qkv_conventional}}
        \hfil
        \subfloat[Separate Non-linear]{
            \includegraphics[width=0.24\textwidth]{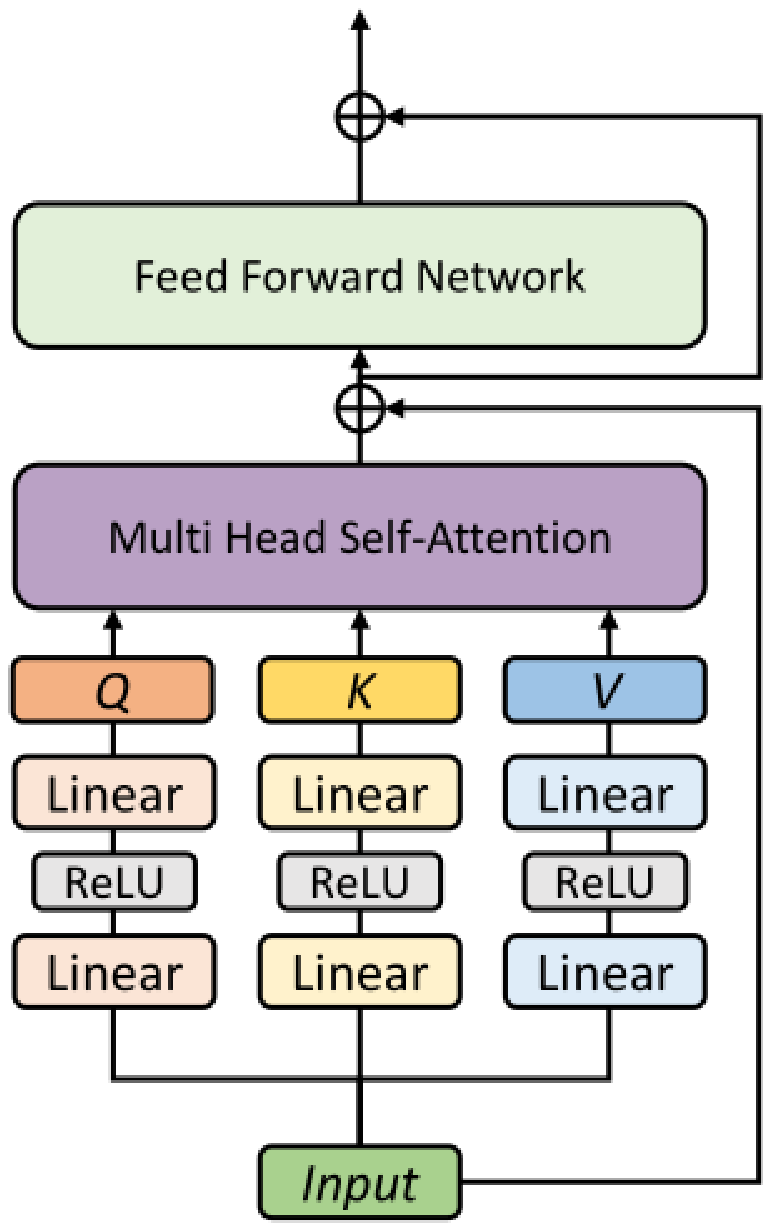}%
            \label{fig_qkv_test1}}
        \hfil
        \subfloat[Partially-Shared Non-linear]{
            \includegraphics[width=0.24\textwidth]{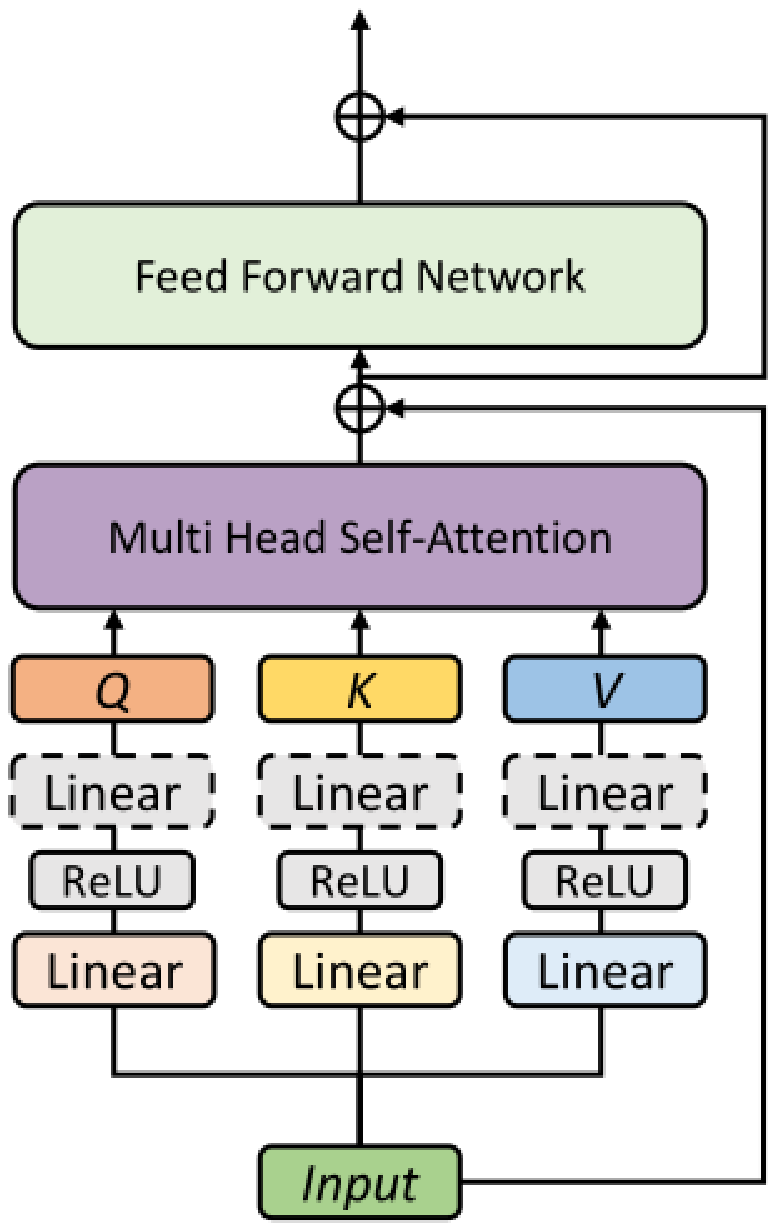}%
            \label{fig_qkv_test2}}
        \hfil
        \subfloat[Fully-Shared Non-linear]{
            \includegraphics[width=0.24\textwidth]{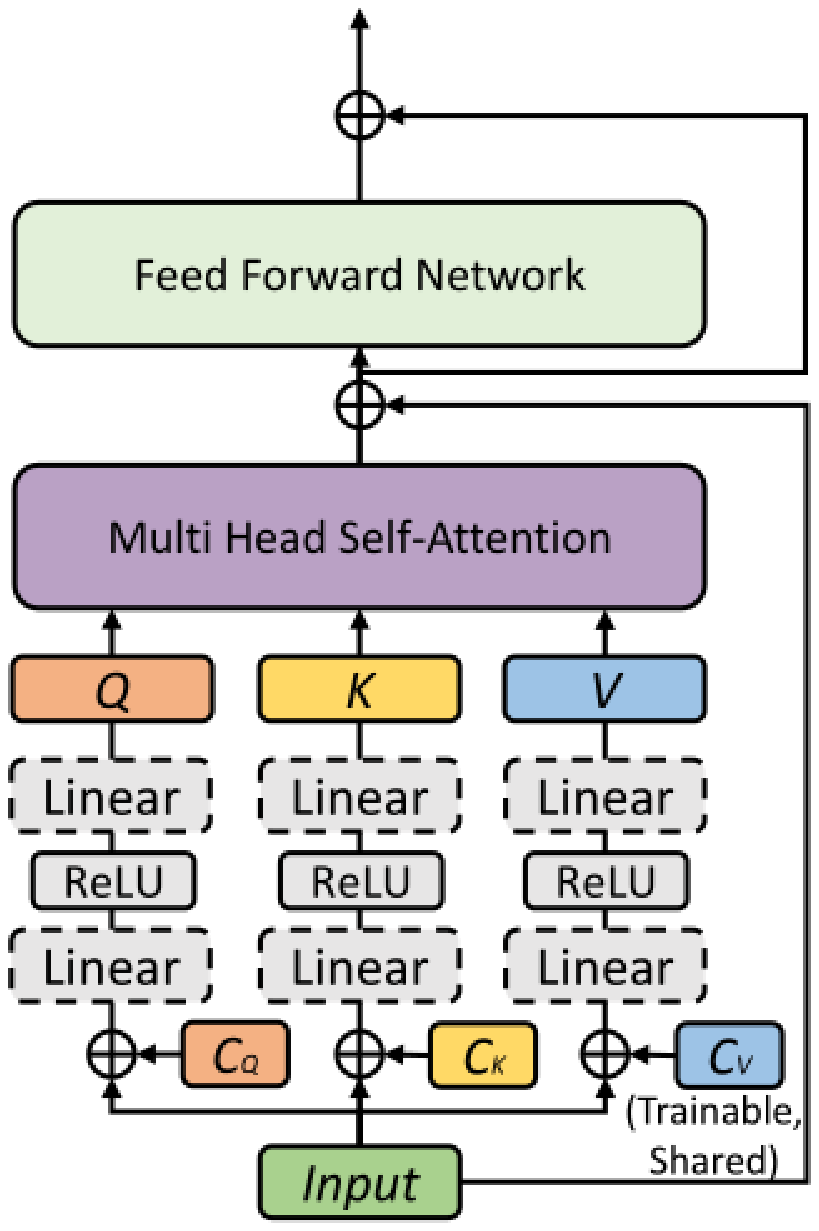}%
            \label{fig_qkv_proposed}}}
    \caption{Four types of $Q, K$, and $V$ embedding methods. (a) The figure depicts conventional embedding used in the standard transformer encoder. (b) Separate non-linear embedding utilizes a two-layer embedding, including non-linearity (ReLU). (c) Partially-shared non-linear embedding includes a weight-sharing linear layer. The weight-sharing layer is depicted as the gray dotted line block, and it follows after the separated linear layer and ReLU. (d) Fully-shared non-linear embedding includes two weight-sharing linear layers, also depicted as a gray dotted line, with ReLU. In this case, trainable codes of $Q, K, V$ (${C}_{q}, {C}_{k}, {C}_{v}$) are concatenated to the input tokens.}
    \vspace{-3mm}
    \label{fig_qkv}
\end{figure*}

\subsection{Self-attention}
Conventional transformers adopt self-attention as the core operation of the network~\cite{Vaswani2017-yi, dosovitskiy2020image}.
In the transformer, the input token $X$ is projected by the linear projection layer $W_{q}$, $W_{k}$, and $W_{v}$ to embed $Q$, $K$, and $V$ vectors, respectively.
Then, the dot product $V$ with the attention matrix, which is obtained by the scaled dot product of $Q$ and $K$.
This self-attention operation can be represented as:
\begin{align}
    SA(Q,K,V)=Softmax\left(\frac{Q\cdot{K^{T}}}{\sqrt{d_{k}}}\right)\cdot{V},
\end{align}
where $SA(Q,K,V)$ is the output of self-attention and $Q, K, V$ are processed by the following operations:
\begin{align}
\begin{split}
    Q=X\cdot{W_{q}}, \\
    K=X\cdot{W_{k}}, \\
    V=X\cdot{W_{v}},
\end{split}
\label{eq:org}
\end{align}
on $X\in{R^{N\times d}}$, $W_{q}\in{R^{d\times d_{q}}}$, $W_{k}\in{R^{d\times d_{k}}}$, $W_{v}\in{R^{d\times d_{v}}}$, where $N$ is the number of tokens, and $d$ is the token dimension.
Self-attention is usually conducted in a multi-headed manner.
When the number of heads is $h$, $d_{q}=d_{k}=d_{v}=d/h$.

\subsection{Cross-covariance Attention}
Cross-covariance attention (XCA) is a modified self-attention mechanism that can reduce the computational complexity from $O(N^{2}d)$ to $O(Nd^{2})$.
XCiT~\cite{El-Nouby2021-ms}, which is a variant of ViT, suggested the use of XCA instead of standard self-attention operation, which is widely used in transformer networks, demonstrating its SOTA performance in CV tasks, for example, ImageNet classification, self-supervised learning, object detection, and semantic segmentation.
They conducted feature dimension self-attention instead of token dimension self-attention, which is used in standard transformer networks.
By simply transposing $Q, K, V$ and reversing the order of the dot product, the computational complexity is reduced from quadratic to linear with the number of tokens, $N$.
This transposed feature dimension self-attention can be represented as below:
\begin{align}
    XCA(Q,K,V)=V\cdot{Softmax\left(\frac{\hat{K}^{T}\cdot{\hat{Q}}}{\tau}\right)},
\label{eq_xca}
\end{align}
where $\hat{Q}, \hat{K}$ are the $l2$-normalized $Q, K$ vector, and $\tau$ is the temperature scaling parameter.
This is orthogonal to our research, which proposes to modify the $Q, K, V$ embedding procedure.
We adopted XCiT as a baseline model to verify the performance of the proposed method.

\section{Proposed method}
As mentioned in Eq. \ref{eq:org}, the conventional embedding method uses a linear layer for each $Q$, $K$, and $V$. The layer is not shared across the embedding spaces of $Q$, $K$, and $V$.
In this paper, we propose three types of new embedding techniques to improve the performance of ViT, as shown in Fig.~\ref{fig_qkv}.

\subsection{Separate Non-linear Embedding (SNE)}
In contrast to the original ViT-based models, we apply non-linear transformations to extract $Q$, $K$, and $V$, respectively, as follows:
\begin{align}
\begin{split}
    Q=\sigma(X\cdot{W_{q}^{(1)}})\cdot{W_{q}^{(2)}}, \\
    K=\sigma(X\cdot{W_{k}^{(1)}})\cdot{W_{k}^{(2)}}, \\
    V=\sigma(X\cdot{W_{v}^{(1)}})\cdot{W_{v}^{(2)}},
\end{split}
\end{align}
where $W_q^{(1)}\in{R^{d\times d_q}}$ and $W_q^{(2)}\in{R^{d_q\times d}}$ represent the weight parameters of the first and second fully connected layers, respectively, and the layers encode the input token as $Q$. $W_k^{(1)}\in{R^{d\times d_k}}$ and $W_k^{(2)}\in{R^{d_k\times d}}$ correspond to $K$, and $W_v^{(1)}\in{R^{d\times d_v}}$ and $W_v^{(2)}\in{R^{d_v\times d}}$ extract $V$. 
$\sigma$ is an activation function. Based on the obtained $Q$, $K$, and $V$, Eq. \ref{eq_xca} is computed for self-attention.

The layers for the SNE consist of two fully connected layers with an activation function (ReLU) to conduct a non-linear transformation of the input tokens.
The non-linear embedding approach may have some advantages. This could increase the total number of non-linearities of the model. Under the limited number of parameters, the increased number of non-linearities could have a positive effect on improving the generalization. Furthermore, it can expand the search space to find new combinations of $Q$, $K$, and $V$.

\subsection{Partially-Shared Non-linear Embedding (P-SNE)}
P-SNE shares a layer from two fully connected layers in the SNE model. There are two options for which the layer will be selected (i.e., first or second layers). Sharing the first layer is similar to the linear embedding originally used in the ViT model. The shared first layer produces the same output because the input values are also shared. Consequently, we chose the second layer to be shared among $Q$, $K$, and $V$.

The shared layer linearly transforms each activation value extracted from the first non-linear layer. With the shared layer, $Q$, $K$, and $V$ can share knowledge on how to build each token. In addition, separate layers of original or SNE structures might have a potential issue, but the shared layer could prevent this issue. In the separate layers, even if one of the layers responsible for $Q$, $K$, and $V$ extraction does not learn well, there might be no problem in minimizing the training loss. This means that the network will be trained well even if one of the three separate layers for $Q$, $K$, and $V$ is not updated.

Finally, $Q$, $K$, and $V$ in the P-SNE, are extracted as follows:
\begin{align}
\begin{split}
    Q=\sigma(X\cdot{W_{q}^{(1)}})\cdot{W_s^{(2)}}, \\
    K=\sigma(X\cdot{W_{k}^{(1)}})\cdot{W_s^{(2)}}, \\
    V=\sigma(X\cdot{W_{v}^{(1)}})\cdot{W_s^{(2)}},
\end{split}
\end{align}
where $W_s^{(2)}\in{R^{d_s\times d}}$ denotes the weight parameters of the shared second layer when $d_s=d_q=d_k=d_v$. We replace the weight parameters $W_q^{(2)}, W_k^{(2)}, W_v^{(2)}$ of the second layer of the SNE with $W_s^{(2)}$.

\subsection{Fully-Shared Non-linear Embedding (F-SNE)}
For F-SNE, we first integrate $Q, K, V$ projection layers $W^{(i)}_{q}, W^{(i)}_{k}, W^{(i)}_{v}$ into the shared projection layers $W^{(i)}_{s}$ for $i=\{1, 2\}$.
Instead of $Q, K, V$ projection layers that separately transform input embedding token $X$ into $Q, K, V$ vectors, the shared projection layers transform the input embedding token $X$ into $Q, K, V$ using the shared weight.
It can guide the projection of $Q, K$, and $V$ towards the same manifold.

However, we could only conduct a single transformation using the shared projection layers, which makes it impossible to embed different vectors simultaneously.
To handle this problem, we add codes ${C}_{q}, {C}_{k}, {C}_{v}$ to reflect separate $Q, K$, and $V$ embedding.
${C}_{q}, {C}_{k}, {C}_{v}$ are trainable vectors that are concatenated to the input token $X$ before passing the shared projection layers $W_{s}$.
Additionally, the same ${C}_{q}, {C}_{k}, {C}_{v}$ are shared among all the encoders in the transformer.
By this sharing, codes will converge at the optimal semantic representation of $Q, K$, and $V$ that exists consistently regardless of the encoder.

$Q, K, V$ embedding by the proposed shared projection layers with codes ${C}_{q}, {C}_{k}, {C}_{v}$ can be comprehensively represented as follows:
\begin{align}
\begin{split}
    Q=\sigma((X\oplus{C_{q}})\cdot{W_s^{(1)}})\cdot{W_s^{(2)}}, \\
    K=\sigma((X\oplus{C_{k}})\cdot{W_s^{(1)}})\cdot{W_s^{(2)}}, \\
    V=\sigma((X\oplus{C_{v}})\cdot{W_s^{(1)}})\cdot{W_s^{(2)}}, 
\end{split}
\label{eq_proposed}
\end{align}
on $X\in{R^{N\times d}}$ and ${C}_{q}, {C}_{k}, {C}_{v}\in{R^{N\times c}}$, where $\oplus$ denotes vector concatenation and $c$ is an arbitrary code size.
Here, ${C}_{q}, {C}_{k}, {C}_{v}\in{R^{1\times c}}$, but they are repeated $N$ times to match the dimension.
The codes ${C}_{q}, {C}_{k}$, and ${C}_{v}$ are computed to minimize the loss function as follows:
\begin{equation}
    \hat{{C}}_q, \hat{{C}}_k, \hat{{C}}_v=\min_{{C}_q, {C}_k, {C}_v} \mathcal{L}(D;\theta, {C}_q, {C}_k, {C}_v) + \lambda \mathcal{R}(\theta),
\label{code_update}
\end{equation}
where $\mathcal{L}$ is a loss function used in the transformer (i.e., cross-entropy loss). $\theta$ denotes the total parameters of the transformer network, and $D$ is the training set. $\mathcal{R}$ represents a regularizer (i.e., weight decay), and $\lambda$ is a parameter that controls the strength of the regularizer.

\section{Experiments and Results}
We evaluated the effect of the new $Q, K$, and $V$ embedding methods on image classification tasks.
We first evaluated the performance of the proposed structures with the ImageNet-1k dataset and then used those models to transfer to the other datasets (i.e., CIFAR-10, CIFAR-100, Stanford Cars, and STL-10).
Additionally, we conducted brief experiments using distillation when we evaluated the performance of the models with the ImageNet-1k dataset.

All the experiments for our methods were conducted using XCiT-Nano and XCiT-Tiny models, which were introduced in \cite{El-Nouby2021-ms} as the baseline models.
In the following sections, we use the brief notations XCiT-N12 and XCiT-T12 to describe the XCiT-Nano and XCiT-Tiny models with 12 repeated encoders.
While the input image size of the models was fixed at 224$\times$224, we trained the models using a batch size of 4,096 for XCiT-N12-based models and 2,816 for XCiT-T12-based models.
For ImageNet training, we trained the model for 400 epochs with an initial learning rate of $5\times 10^{-4}$.
For transfer learning, we used an initial learning rate of $5\times 10^{-5}$ for 1,000 epochs of training.
We followed other experimental setups of the original XCiT.
Overall experiments were conducted on NVIDIA DGX A100 (8 GPUs).

For the $Q, K$, and $V$ projection layers, we used the output dimensions depicted in Table~\ref{tab_dimension} according to the variants.
Additionally, in the case of the XCiT-N12 (F-SNE, $c$) model, the shared projection layers have an input dimension of $(128+c)$ to take the input concatenated with the code, and the XCiT-T12 (F-SNE, $c$) model has shared projection layers with an input dimension of $(192+c)$.
Code is a parameter vector of arbitrary code size (\eg, c = 8, 16, 32, 64) defined outside the encoder to be shared among the encoders.
For a fair comparison, we compare the number of parameters as well as the performance in the following sections. 
In the case of the F-SNE models, codes were also included in the model parameters.
\begin{table}[h]
\small
    \caption{Corresponding output dimensions of Q, K, V embedding layers to each model variant.}
    \label{tab_dimension}
    \centering
    \begin{tabular}{c|c|c|c|c}
        \hline
        \multirow{2}{*}{Model} & \multicolumn{2}{c}{Nano} & \multicolumn{2}{|c}{Tiny}\\
        \cline{2-5}
             & $d$ & $d_{q}, d_{k}, d_{v}$ & $d$ & $d_{q}, d_{k}, d_{v}$ \\
        \hline
        \hline
        XCiT \cite{El-Nouby2021-ms} & 128 & 128 & 192 & 192\\
        XCiT (SNE) & 128 & 64 & 192 & 96\\
        XCiT (P-SNE) & 128 & 96 & 192 & 144\\
        XCiT (F-SNE, $c$) & 128 & 128 & 192 & 192\\
        XCiT (F-SNE, 8*) & 128 & 186 & 192 & 282\\
        XCiT (F-SNE, 16*) & 128 & 182 & 192 & 276\\
        \hline
    \end{tabular}
\end{table}

\begin{table}[h]
\small
    \caption{Evaluation on ImageNet classification task with $<$10M parameters. In Type, `C' denotes CNN architecture, and `T' denotes the transformer architecture. Gray color represents our methods.}
    \vspace{-2mm}
    \label{tab_imagenet}
    \centering
    \begin{tabular}{c|c|c|c}
        \hline
        Model  & param \# & Type & Top-1 (\%)\\
        \hline
        \hline
        Mobile-Former-26M \cite{chen2021mobile} & 3.2M & C+T & 64.0 \\ %
        Mobile-Former-52M \cite{chen2021mobile} & 3.5M & C+T & 68.7 \\
        MobileNetV3-S \cite{Howard2019-tj} & 2.5M & C & 67.4 \\
        MobileNetV3-S/1.25x \cite{Howard2019-tj} & 3.6M & C & 70.4 \\
        \hline
        XCiT-N12 (baseline)\cite{El-Nouby2021-ms} & 3.1M & T & 69.9 \\
        \rowcolor{Gray}
        XCiT-N12 (SNE)& 3.1M & T & {70.9} (+1.0)\\
        \rowcolor{Gray}
        XCiT-N12 (P-SNE) & 3.1M & T & \textbf{71.4} (+1.5)\\
        \rowcolor{Gray}
        XCiT-N12 (F-SNE, 8) & 2.9M & T & {70.6} (+0.7)\\
        \rowcolor{Gray}
        XCiT-N12 (F-SNE, 8*) & 3.1M & T & {70.8} (+0.9)\\
        \rowcolor{Gray}
        XCiT-N12 (F-SNE, 16) & 2.9M & T & {70.2} (+0.3)\\
        \rowcolor{Gray}
        XCiT-N12 (F-SNE, 16*) & 3.1M & T & {70.6} (+0.7)\\
        \rowcolor{Gray}
        XCiT-N12 (F-SNE, 32) & 2.9M & T & {70.3} (+0.4)\\
        \rowcolor{Gray}
        XCiT-N12 (F-SNE, 64) & 2.9M & T & {70.2} (+0.3)\\
        \hline
        \hline
        ViT-Ti \cite{dosovitskiy2020image} & 5.7M & T & 68.7 \\
        T2T-ViT-7 \cite{yuan2021tokens} & 4.3M & T & 71.7 \\
        DeiT-Ti \cite{Touvron2020-io} & 5.7M & T & 72.2 \\
        Mobile-Former-96M \cite{chen2021mobile} & 4.6M & C+T & 72.8 \\
        LocalViT-T2T \cite{li2021localvit} & 4.3M & C+T & 72.5 \\
        PiT-Ti \cite{heo2021rethinking} & 4.9M & C+T & 73.0 \\ %
        ConViT-Ti \cite{d2021convit} & 5.7M & C+T & 73.1 \\
        ConT-Ti \cite{yan2021contnet} & 5.8M & C+T & 74.9 \\
        LocalViT-T \cite{li2021localvit} & 5.9M & C+T & 74.8 \\
        ViTAE-T \cite{xu2021vitae} & 4.8M & C+T & 75.3 \\
        EfficientNet-B0 \cite{Tan2019-zx} & 5.3M & C & 76.3 \\
        CeiT-T \cite{yuan2021incorporating} & 6.4M & C+T & 76.4 \\
        T2T-ViT-12 \cite{yuan2021tokens} & 6.9M & T & 76.5 \\
        ConT-S \cite{yan2021contnet} & 10.1M & C+T & 76.5 \\
        CoaT-Lite Tiny \cite{xu2021co} & 5.7M & C+T & 76.6 \\
        Swin-1G \cite{liu2021swin, chen2021mobile} & 7.3M & T & 77.3 \\
        \hline
        XCiT-T12 (baseline) \cite{El-Nouby2021-ms} & 6.7M & T & 77.1 \\
        \rowcolor{Gray}
        XCiT-T12 (SNE) & 6.7M & T & {77.6} (+0.5) \\
        \rowcolor{Gray}
        XCiT-T12 (P-SNE) & 6.7M & T & \textbf{77.8} (+0.7) \\
        \rowcolor{Gray}
        XCiT-T12 (F-SNE, 8) & 6.3M & T & {76.8} (-0.3) \\
        \rowcolor{Gray}
        XCiT-T12 (F-SNE, 8*) & 6.7M & T & {77.7} (+0.6)\\
        \rowcolor{Gray}
        XCiT-T12 (F-SNE, 16) & 6.3M & T & {77.2} (+0.1) \\
        \rowcolor{Gray}
        XCiT-T12 (F-SNE, 16*) & 6.7M & T & {77.4} (+0.3)\\
        \rowcolor{Gray}
        XCiT-T12 (F-SNE, 32) & 6.3M & T & {77.1} (0.0) \\
        \rowcolor{Gray}
        XCiT-T12 (F-SNE, 64) & 6.3M & T & {77.1} (0.0) \\
        \hline
    \end{tabular}
\end{table}

\subsection{ImageNet Classification}
Table~\ref{tab_imagenet} shows the ImageNet classification results to compare other models with our proposed method.
Top-1 accuracies from other models noted in the table came directly from each model's publications.
In this section, we used ImageNet-1k, which consists of 1.28M training images and 50,000 validation images with 1,000 labels.
To demonstrate the performance of the proposed method, we selected XCiT as a baseline model and modified $Q, K$, and $V$ projection layers corresponding to each model variant.

As shown in Table~\ref{tab_imagenet}, the P-SNE model achieved the best score for the below 4M parameters constraints, improving the accuracy by 1.5\% from the baseline.
Even in the case of the below 10M parameters constraints, the P-SNE model surpassed the previous SOTA model, which recorded 77.3\% top-1 accuracy while improving the accuracy by 0.5\% from it.
The starred (*) model is an additional experiment to compare the performance of our method using the same number of parameters as the original XCiT model.
This shows that the performance can be improved by adding parameters to the embedding layers as much as the capacity saved by fully sharing the embedding layers.
For example, the XCiT-T12 (F-SNE, 8*) model scored 77.7\%, which is close to the best accuracy of 77.8\%.
Moreover, the non-linear embedding method (SNE) also improves the classification rates on the ImageNet dataset, compared to the baseline model which uses the linear embedding method.

\subsection{Distillation}
We evaluated the performances on ImageNet classification task using distillation technique as well.
In this section, we used RegNetY-16GF~\cite{radosavovic2020designing} as a teacher model to conduct hard distillation as proposed in \cite{Touvron2020-io}.
Similar to the previous experimental results in \cite{Touvron2020-io, El-Nouby2021-ms}, distillation could improve the performance of each model.
Additionally, shared non-linear embedding methods improve the performance of the baseline model, although a separate non-linear embedding method decreases the performance.
This reminds us of the better properties of the shared embedding methods, which is consistent with the results shown in the previous section.
These results are organized in Table~\ref{tab_imagenet_distillation}.

\begin{table}[h!]
\small
    \caption{Evaluation on ImageNet classification with distillation. Gray color represents our methods.}
    \vspace{-3mm}
    \label{tab_imagenet_distillation}
    \centering
    \begin{tabular}{c|c|c}
        \hline
        Model  & param \# & Top-1 (\%)\\
        \hline
        \hline
        XCiT-N12 (baseline)\S \cite{El-Nouby2021-ms} & 3.1M & 71.7 \\
        \rowcolor{Gray}
        XCiT-N12 (SNE)\S & 3.1M & {71.5} (-0.2)\\
        \rowcolor{Gray}
        XCiT-N12 (P-SNE)\S & 3.1M & {71.9} (+0.2)\\
        \rowcolor{Gray}
        XCiT-N12 (F-SNE, 8)\S & 2.9M & {71.7} (0.0)\\
        \rowcolor{Gray}
        XCiT-N12 (F-SNE, 8*)\S & 3.1M & \textbf{72.4} (+0.7)\\
        \rowcolor{Gray}
        XCiT-N12 (F-SNE, 16)\S & 2.9M & {72.0} (+0.3)\\
        \rowcolor{Gray}
        XCiT-N12 (F-SNE, 16*)\S & 3.1M & {72.2} (+0.5)\\
        \rowcolor{Gray}
        XCiT-N12 (F-SNE, 32)\S & 2.9M & {71.9} (+0.2)\\
        \rowcolor{Gray}
        XCiT-N12 (F-SNE, 64)\S & 2.9M & {71.8} (+0.1)\\
        \hline
    \end{tabular}
\end{table}

\begin{table*}[h!]
\small
    \centering
    \caption{Evaluation on transfer learning. All models were pre-trained using the ImageNet-1k dataset. \dag~indicates the results are obtained from the paper. \ddag~indicates that the experiment cannot be performed due to the lack of source code. Gray color represents our methods.}
    \begin{tabular}{c|c|c|c|c|c|c}
    \hline
    Model & param \# & CIFAR-10 (\%) & CIFAR-100 (\%) & Cars (\%) & STL-10 (\%) & Average (\%) \\
    \hline
    \hline
    MobileNetV3-S \cite{Howard2019-tj} & 2.9M & 96.6 & 82.3 & 79.8 & 94.3 & 88.3\\
    MobileNetV3-L/0.75x \cite{Howard2019-tj} & 4.0M & 97.5 & 84.9 & 87.6 & 96.5 & 91.6\\
    RegNetY-200MF \cite{radosavovic2020designing} & 3.2M & 97.4 & 84.2 & 86.3 & 96.2 & 91.0\\
    RegNetY-400MF \cite{radosavovic2020designing} & 4.3M & 97.8 & 84.6 & 90.7 & 97.0 & 92.5\\
    EfficientNet-B0 \cite{Tan2019-zx} & 5.3M & 98.1$^\dag$ & 88.1$^\dag$ & 90.8 & 98.1 & 93.8 \\
    ViTAE-T \cite{xu2021vitae} & 4.8M & 97.3$^\dag$ & 86.0$^\dag$ & 89.5$^\dag$ & \_$^\ddag$ & \_ \\
    CeiT-T \cite{yuan2021incorporating} & 6.4M & 98.5$^\dag$ & \textbf{88.4}$^\dag$ & 90.5$^\dag$ & 98.3 & 93.9 \\
    \hline
    XCiT-N12 (baseline) \cite{El-Nouby2021-ms} & 3.1M & 98.0 & 85.5 & 87.9 & 97.5 & 92.2 \\
    \rowcolor{Gray}
    XCiT-N12 (SNE) & 3.1M & {98.4} (+0.4) & 86.6 (+1.1) & 88.9 (+1.0) & 97.8 (+0.3) & 92.9 (+0.7)\\
    \rowcolor{Gray}
    XCiT-N12 (P-SNE) & 3.1M & 98.2 (+0.2) & 86.5 (+1.0) & 88.9 (+1.0) & {97.9} (+0.4) & 92.9 (+0.7) \\
    \rowcolor{Gray}
    XCiT-N12 (F-SNE, 8) & 2.9M & 98.1 (+0.1) & 86.2 (+0.7) & 89.0 (+1.1) & 97.8 (+0.3) & 92.8 (+0.6)\\
    \rowcolor{Gray}
    XCiT-N12 (F-SNE, 8*) & 3.1M & 98.3 (+0.3) & 86.2 (+0.7) & 89.0 (+1.1) & 97.8 (+0.3) & 92.8 (+0.6)\\
    \rowcolor{Gray}
    XCiT-N12 (F-SNE, 16) & 2.9M & 98.2 (+0.2) & 86.3 (+0.8) & 90.2 (+2.3) & 97.8 (+0.3) & 93.1 (+0.9)\\
    \rowcolor{Gray}
    XCiT-N12 (F-SNE, 16*) & 3.1M & 98.3 (+0.3) & 86.3 (+0.8) & 90.5 (+2.6) & {97.9} (+0.4) & 93.2 (+1.0)\\
    \rowcolor{Gray}
    XCiT-N12 (F-SNE, 32) & 2.9M & 98.1 (+0.1) & {86.8} (+1.3) & {90.6} (+2.7) & 97.8 (+0.3) & {93.3} (+1.1)\\
    \rowcolor{Gray}
    XCiT-N12 (F-SNE, 64) & 2.9M & 98.1 (+0.1) & 86.2 (+0.7) & 89.1 (+1.2) & 97.6 (+0.1) & 92.8 (+0.6)\\
    \hline
    XCiT-T12 (baseline) \cite{El-Nouby2021-ms} & 6.7M & 98.5 & 86.7 & \textbf{92.7} & 98.3 & 94.0\\
    \rowcolor{Gray}
    XCiT-T12 (SNE) & 6.7M & 98.6 (+0.1) & 87.2 (+0.5) & 91.8 (-0.9)  & 98.6 (+0.3) & 94.0 (0.0) \\
    \rowcolor{Gray}
    XCiT-T12 (P-SNE) & 6.7M & \textbf{98.7} (+0.2) & 87.2 (+0.5) & 91.8 (-0.9) & 98.6 (+0.3) & 94.1 (+0.1) \\
    \rowcolor{Gray}
    XCiT-T12 (F-SNE, 8) & 6.3M & 98.4 (-0.1) & {88.0} (+1.3) & 92.3 (-0.4) & 98.6 (+0.3) & \textbf{94.3} (+0.3)\\
    \rowcolor{Gray}
    XCiT-T12 (F-SNE, 8*) & 6.7M & 98.5 (0.0) & 87.6 (+0.9) & 92.1 (-0.6) & 98.3 (0.0) & 94.1 (+0.1) \\
    \rowcolor{Gray}
    XCiT-T12 (F-SNE, 16) & 6.3M & 98.5 (0.0) & 87.2 (+0.5) & 92.3 (-0.4) & \textbf{98.7} (+0.4) & 94.2 (+0.2)\\
    \rowcolor{Gray}
    XCiT-T12 (F-SNE, 16*) & 6.3M & 98.6 (+0.1) & 87.8 (+1.1) & 92.3 (-0.4) & 98.6 (+0.3) & \textbf{94.3} (+0.3)\\
    \rowcolor{Gray}
    XCiT-T12 (F-SNE, 32) & 6.3M & 98.5 (0.0) & 87.9 (+1.2) & 92.3 (-0.4) & 98.4 (+0.1) & \textbf{94.3} (+0.3)\\
    \rowcolor{Gray}
    XCiT-T12 (F-SNE, 64) & 6.4M & 98.5 (0.0) & 87.7 (+1.0) & 92.3 (-0.4) & 98.5 (+0.2) & \textbf{94.3} (+0.3)\\    
    \hline
    \end{tabular}
    \label{tab_transfer}
\end{table*}

\subsection{Transfer Learning}
To demonstrate the generalization performance of our method, we conducted transfer learning experiments on CIFAR-10, CIFAR-100~\cite{krizhevsky2009learning}, Stanford Cars~\cite{Krause2013-ui}, and STL-10~\cite{coates2011analysis} datasets as shown in Table~\ref{tab_transfer}. 
CIFAR-10 and CIFAR-100 consist of 50,000 training images and 10,000 test images with 10 and 100 classes, respectively.
The Stanford Cars dataset consists of 8,144 training images and 8,041 test images of 196 classes.
Lastly, we used the STL-10 dataset, which contains 5,000 training images and 8,000 test images from 10 classes.

As shown in the average score of Table~\ref{tab_transfer}, our method generally improves the performance of the baseline model.
In particular, XCiT-N12 (F-SNE, 32) and XCiT-T12 (F-SNE, 8) achieved the best score under each parameter constraint, below 4M and 10M, despite having fewer parameters.
Including these best models, F-SNE models showed mostly better transfer performance compared to other variants on average.
It can be assumed that fully shared embedding may provide better generalization capability than other variants.
However, on the Cars dataset, there were performance drops in the case of transferring XCiT-T12-based models, as opposed to the case of transferring XCiT-N12-based models.
The embedding dimension of XCiT-T12 is larger than that of XCiT-N12, which means that XCiT-T12 has an extra capacity to find a complex mapping function.
It may decrease the contribution of non-linear embeddings.
However, it is still worth using non-linear embedding methods in small models.

\section{Ablation Study and Analysis}
We performed several ablation studies to analyze the proposed embedding structures, such as variation in the number of non-linear layers, comparison of shared and unshared code, code visualization of F-SNE, and code size search.

\subsection{Impact of the Number of Layers}
\begin{figure}[h!]
    \centering\includegraphics[width=\linewidth]{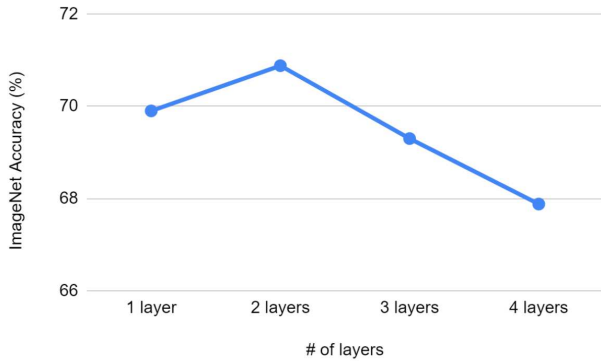}
    \vspace{-7mm}
    \caption{Performance comparison between the number of non-linear layers from 1 to 4. The two layers achieved the best accuracy, while three or four layers showed lower accuracies.}
    \vspace{-3mm}
    \label{fig_layer_number}
\end{figure}

All the proposed methods utilize the two-layer model for non-linear embedding, which is the smallest unit that expresses non-linearity. To analyze the impact of the number of layers, we performed an analysis according to the number of layers. Fig. \ref{fig_layer_number} shows the results based on the XCiT-N12 (SNE) model using the ImageNet dataset. 

The accuracies with three or four layers decreased, but the best accuracy was achieved when two layers were used. Based on the results, we conclude that more than a single non-linearity embedding can hinder the embedding performance in terms of accuracy.  

\subsection{Sharing or Unsharing Codes in F-SNE}
Owing to the code, the input token can be identified in the F-SNE for separately extracting $Q$, $K$, and $V$. In the F-SNE model, codes ${C}_q, {C}_k$, and ${C}_v$ are shared across all embedding modules in the transformer model. This is effective for slightly decreasing the total number of parameters. In addition, as shown in Table \ref{tab_share_unshare}, the performance can be improved. Without sharing, the codes might have different values, which could hinder to find the optimal code vectors for $Q, K$, and $V$. Sharing might be a good prior knowledge to reach the optimal solution.

\begin{table}[H]
\small
    \caption{Comparison between the cases of code sharing and unsharing tested on the ImageNet dataset. The numbers in brackets denote the total number of parameters for each model.}
    \label{tab_share_unshare}
    \centering
    \begin{tabular}{c|c|c}
        \hline
        Model & Sharing & W/o sharing\\
        \hline
        \hline
        XCiT-N12 (F-SNE, 8*) & \textbf{70.82} (2.9M) & {70.28 (3.05M)}\\
        \hline
    \end{tabular}
\end{table}

\subsection{${C}_q, {C}_k$, and ${C}_v$ in Different Tasks}
The optimal code values were initially found on the ImageNet dataset, and then the computed code values were used as the initial codes for the downstream tasks. Even if the pre-trained codes are used, the code values might vary according to the tasks because the codes are updated via a back-propagation process using the downstream task dataset.

\begin{figure}[!ht]
        \subfloat[ImageNet-CIFAR10]{
            \includegraphics[width=0.49\linewidth]{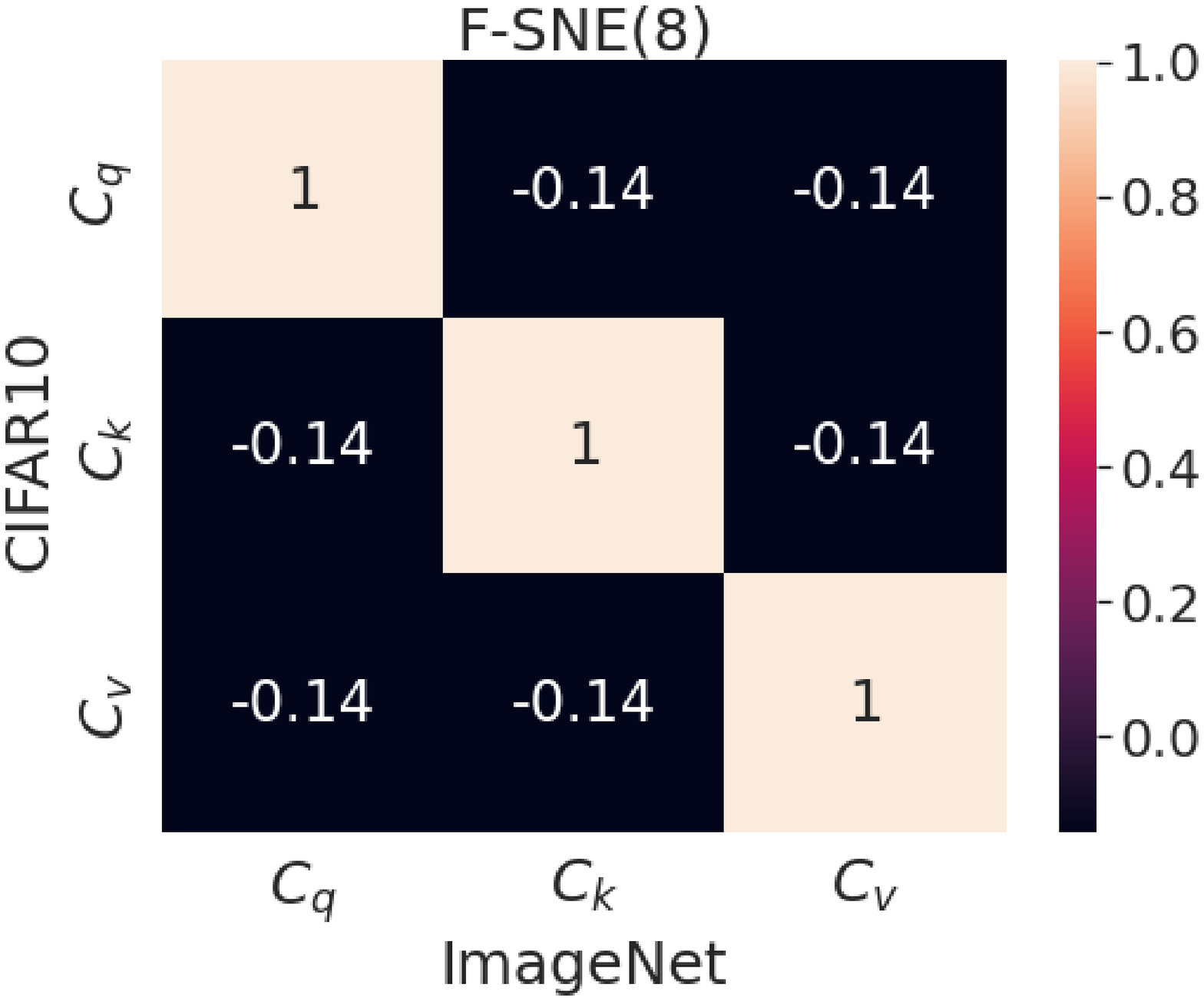}%
            \label{fig_code_imagenet_cifar10}}
        \subfloat[ImageNet-CIFAR100]{
            \includegraphics[width=0.49\linewidth]{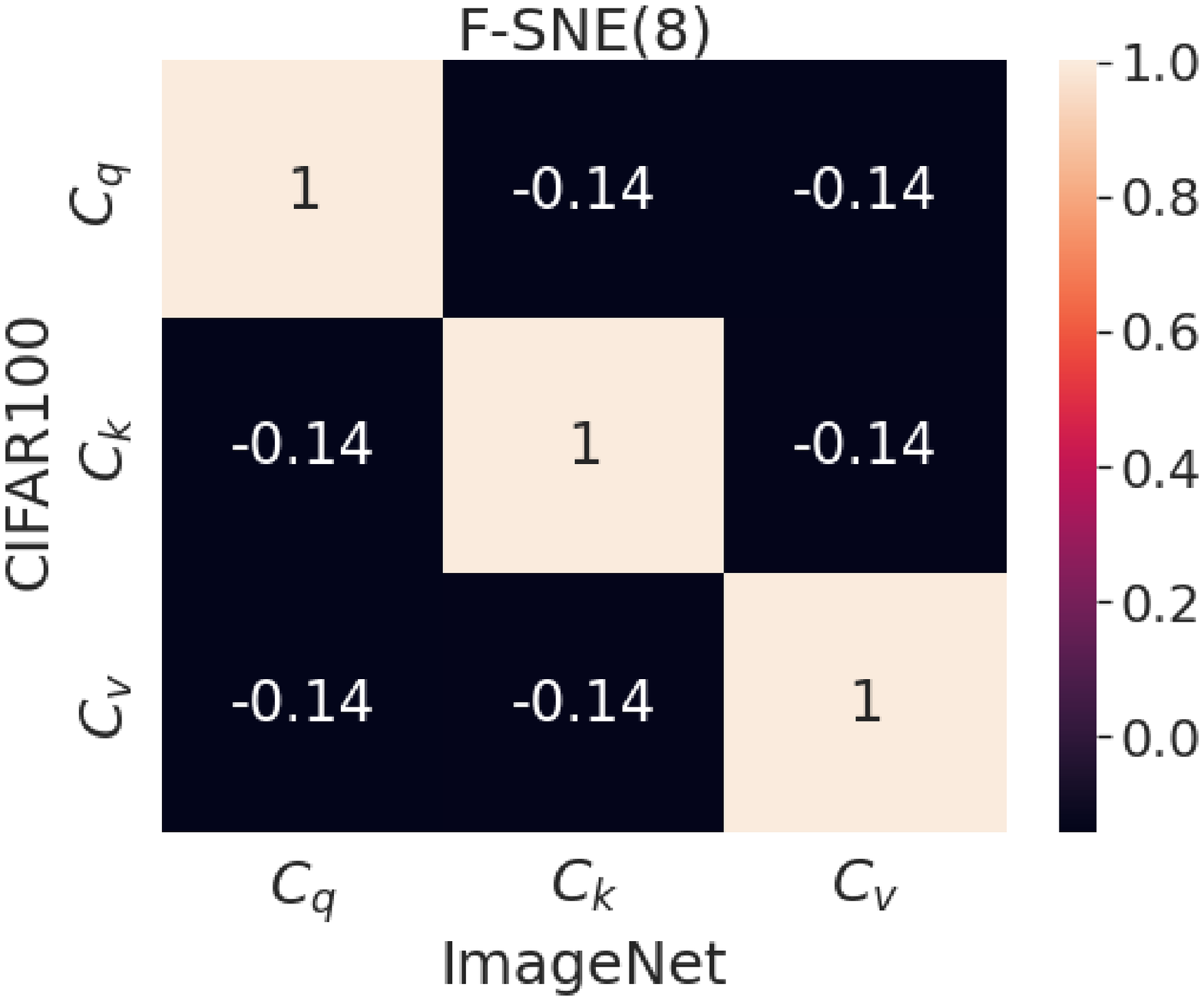}%
            \label{fig_code_imagenet_cifar100}}\\
        \\
        \subfloat[ImageNet-Cars]{
            \includegraphics[width=0.49\linewidth]{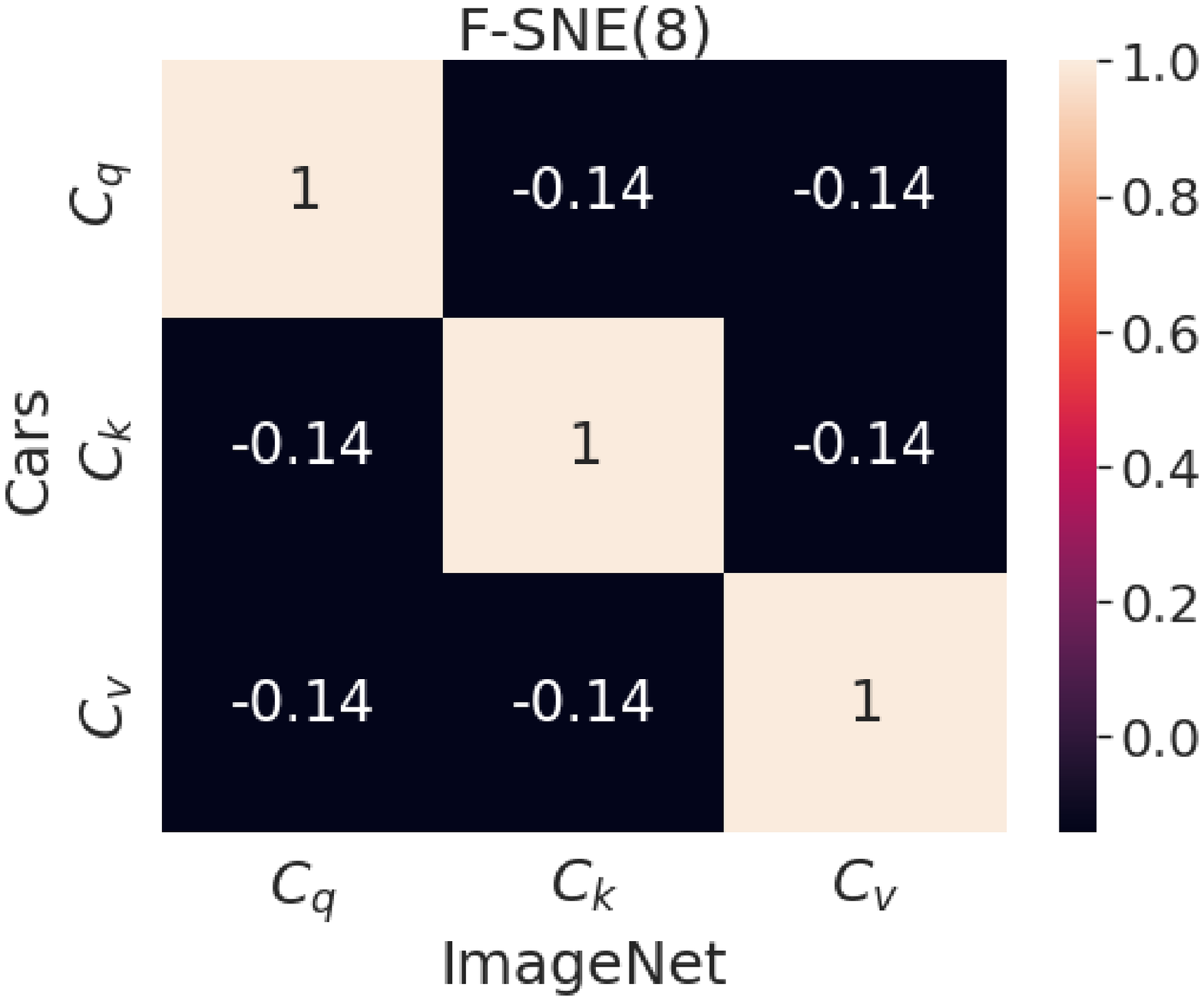}%
            \label{fig_code_imagenet_cars}}
        \subfloat[ImageNet-STL10]{
            \includegraphics[width=0.49\linewidth]{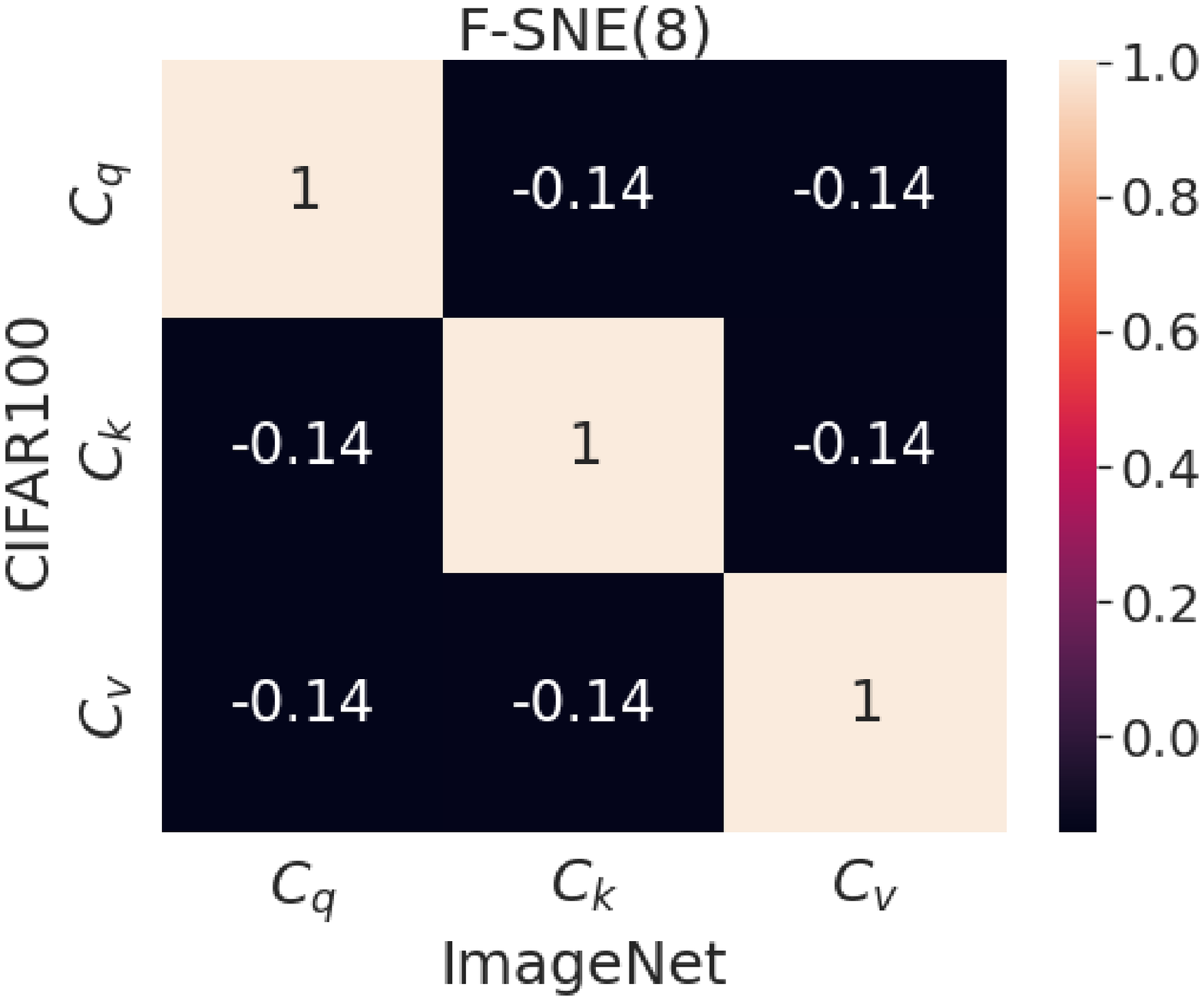}%
            \label{fig_code_imagenet_stl10}}
    \caption{Dot product value of $l2$-normalized ${C}_q, {C}_k$, and ${C}_v$ according to the trained datasets. ${C}_q, {C}_k$, and ${C}_v$ are extracted from the XCiT-N12 (F-SNE, 8) model.}
    \vspace{-4mm}
    \label{fig_code_visualize}
\end{figure}

Figure \ref{fig_code_visualize} shows correlation values of the codes extracted from each downstream task. Interestingly, the correlation matrices are very similar even if the task is changed. 
The diagonal elements give some information on the similarity between the same codes, and non-diagonal elements depict the similarity between different codes.
Non-diagonal elements are close to the zero, and each code of ${C}_q$, ${C}_k$, and ${C}_v$ tends to have the property of orthogonality. These results can be interpreted that ${C}_q$, ${C}_k$, and ${C}_v$ learn their inherent feature to be used as $Q, K$, and $V$, regardless of the task; furthermore, the values of the $l2$-norm for each dataset can be obtained, as presented in Table \ref{tab_code_l2norm}. The codes of ImageNet, Cars, and STL-10 have similar $l2$-norm values, but CIFAR-10 and 100 have different $l2$-norm values. 

\begin{table}[h!]
\small
    \centering
    \caption{$l2$-norm of ${C}_{q}$, ${C}_{k}$, and ${C}_{v}$ extracted from XCiT-N12 (F-SNE, 8) model according to the trained dataset.}
    \begin{tabular}{c|c|c|c}
    \hline
    Dataset & $\norm{C_q}$ & $\norm{C_k}$ & $\norm{C_v}$ \\
    \hline
    \hline
    ImageNet & 8.86 & 8.13 & 8.53 \\
    CIFAR-10 & 9.05 & 8.34 & 8.74 \\
    CIFAR-100 & 9.06 & 8.35 & 8.77 \\
    Cars & 8.84 & 8.13 & 8.53 \\
    STL-10 & 8.86 & 8.14 & 8.56 \\
    \hline
    \end{tabular}
    \label{tab_code_l2norm}
\end{table}

\begin{figure*}[t!]
    \centerline{
        \subfloat[ImageNet-CIFAR10]{
            \includegraphics[width=0.24\textwidth]{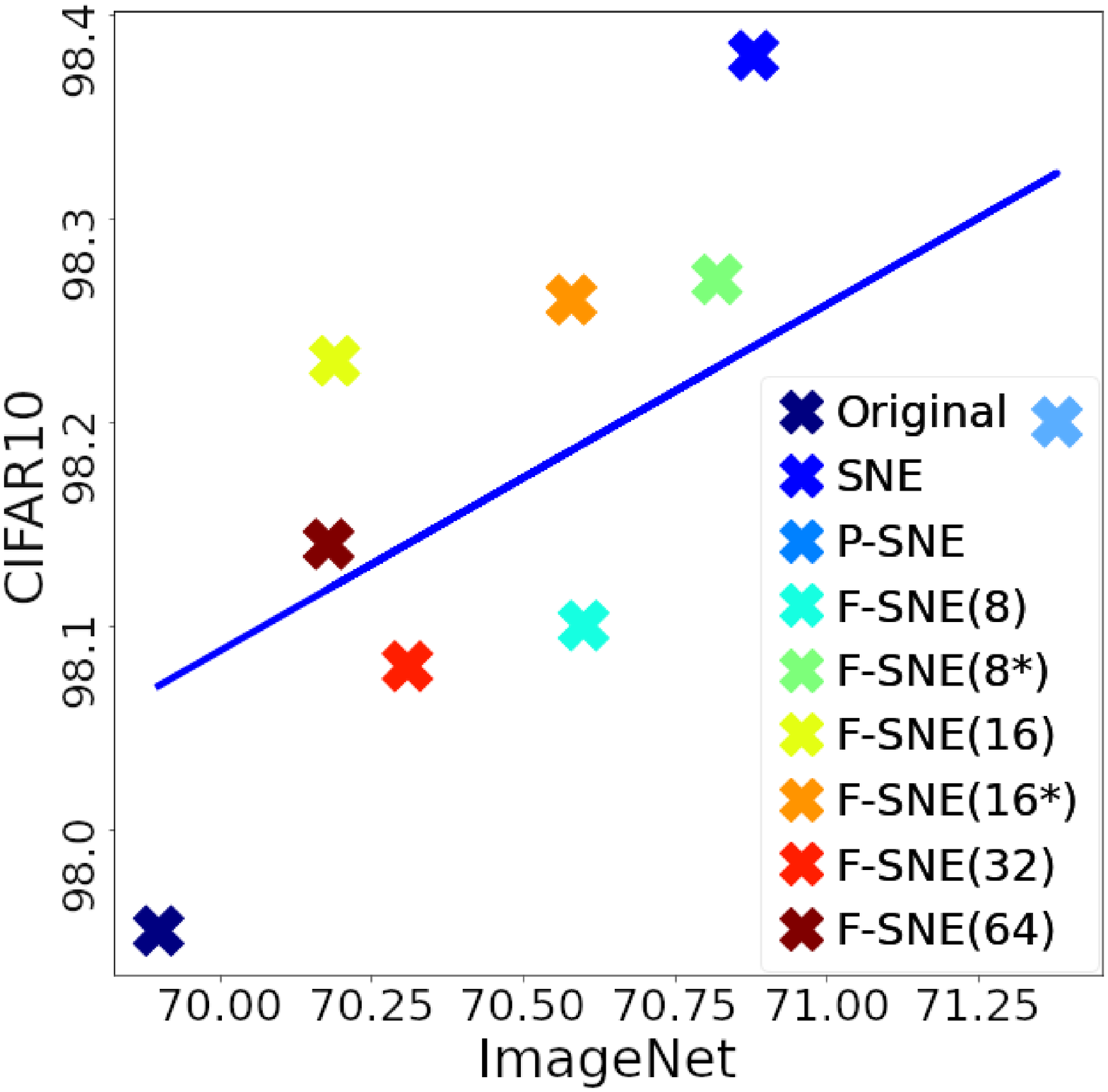}%
            \label{fig_correlation_cifar10}}
        \hfil
        \subfloat[ImageNet-CIFAR100]{
            \includegraphics[width=0.24\textwidth]{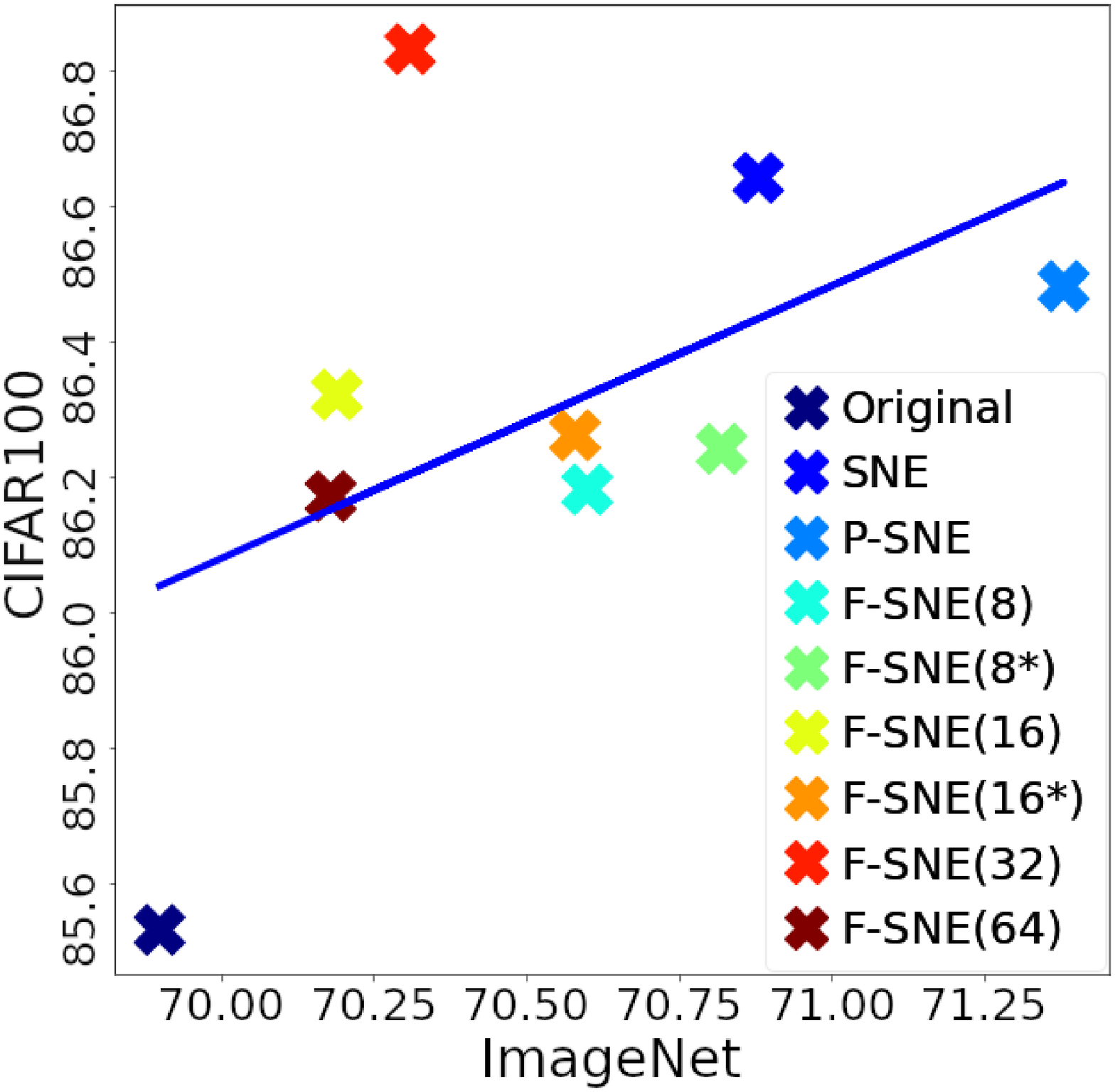}%
            \label{fig_correlation_cifar100}}
        \hfil
        \subfloat[ImageNet-Cars]{
            \includegraphics[width=0.24\textwidth]{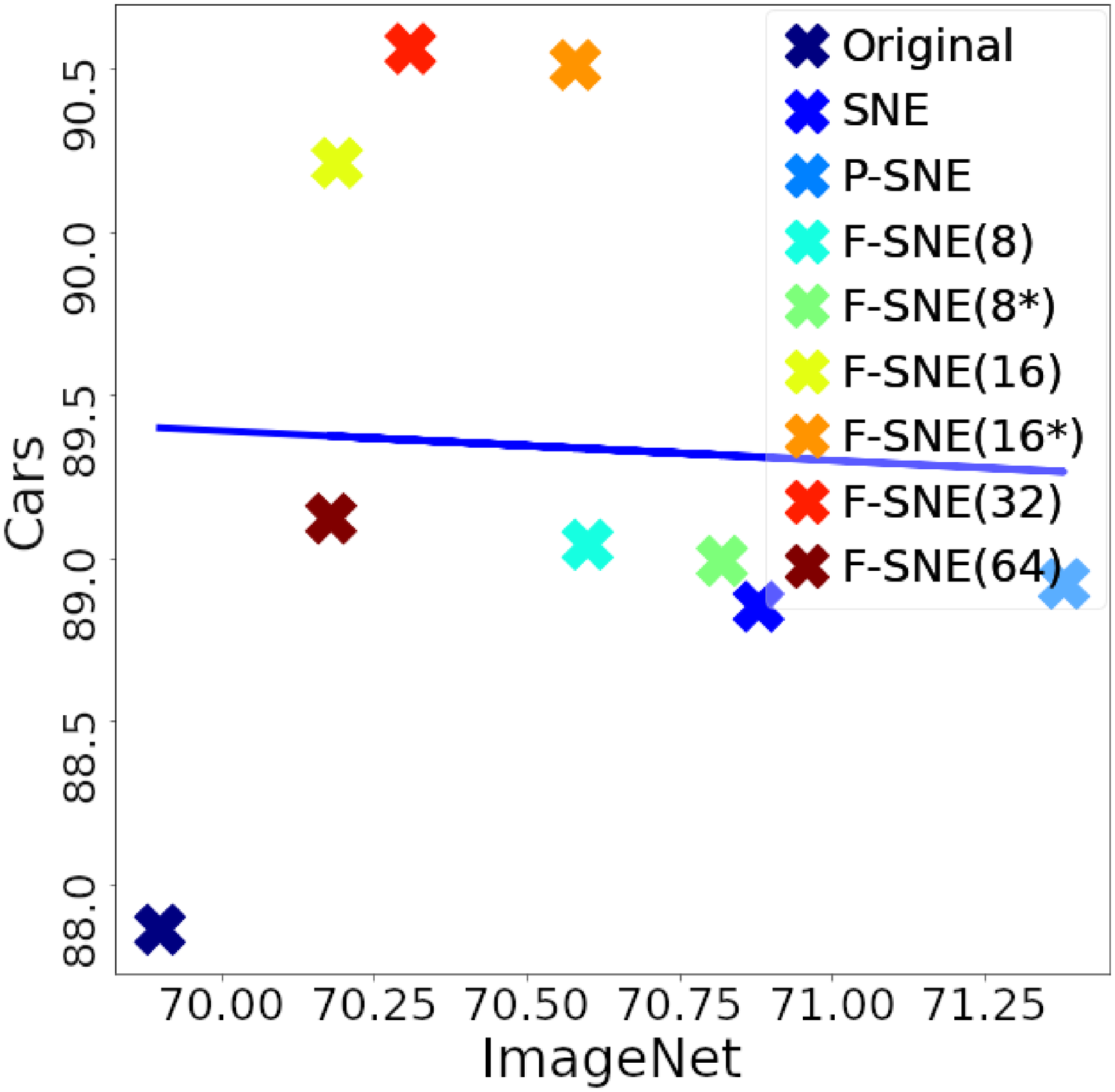}%
            \label{fig_correlation_cars}}
        \hfil
        \subfloat[ImageNet-STL10]{
            \includegraphics[width=0.24\textwidth]{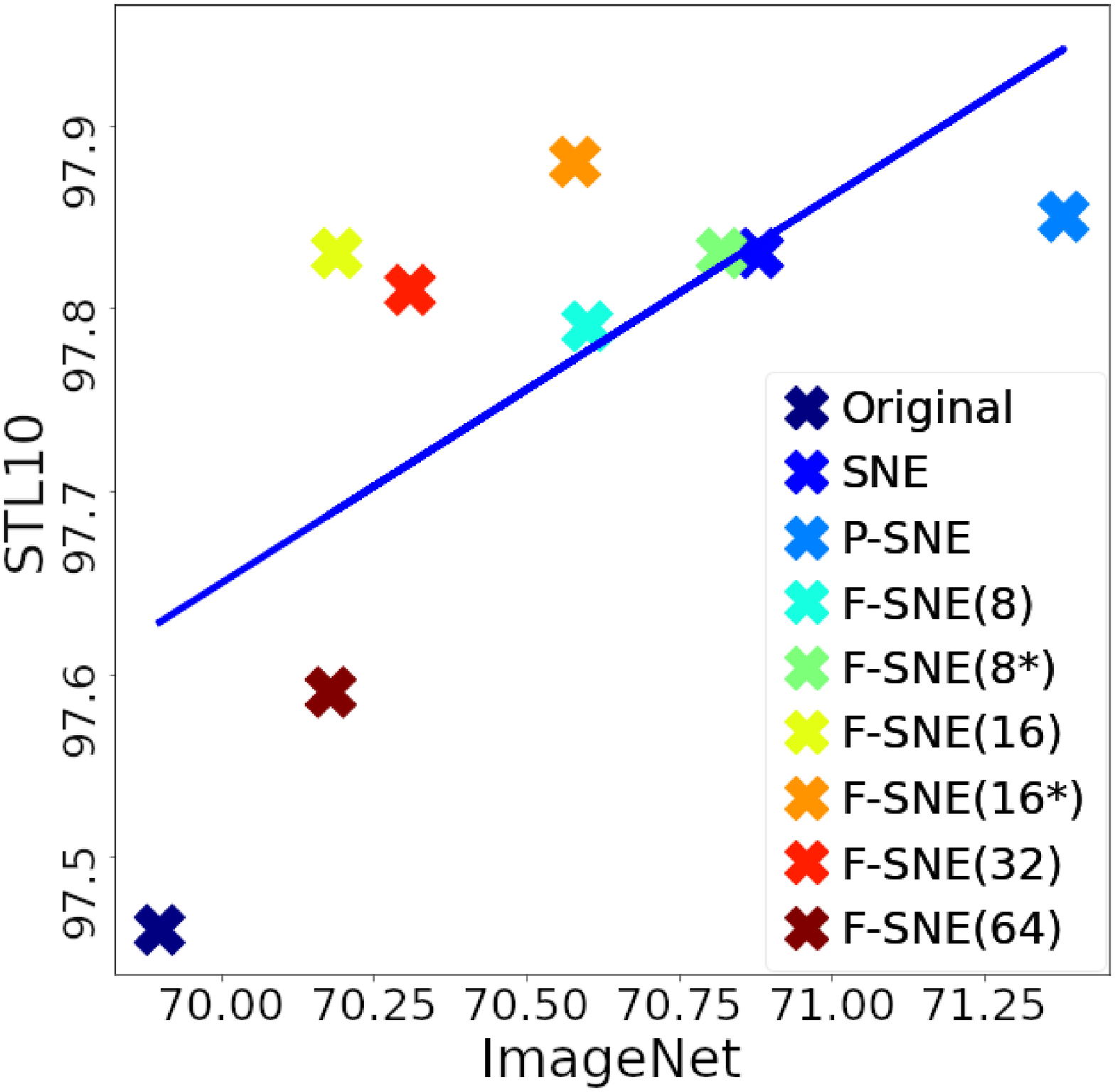}%
            \label{fig_correlation_stl10}}}
    \vspace{-2mm}
    \caption{Correlation plots of pre-training and transfer learning performances. XCiT-N12-based models are used. 'X' markers denote the accuracies of each model, and the blue line denotes the regression slope of the results. In contrast to other correlated plots, the results on the Cars dataset are not correlated with the results on the ImageNet dataset. Best viewed in color.}
    \vspace{-3mm}
    \label{fig_correlation}
\end{figure*}

\subsection{Do Better ImageNet Models Transfer Better?}
In general, better ImageNet models transfer better \cite{kornblith2019better}. An experiment was conducted to verify whether this fact also applies to our cases. As shown in Fig. \ref{fig_correlation}, we could observe correlation between performances on ImageNet and on downstream tasks. The correlation tendency is reflected by the domain difference between the upstream and downstream tasks. In other words, the tasks of the CIFAR and STL datasets are similar to the task of ImageNet, so they show high correlations between two tasks. However, in the Stanford Cars dataset, it is difficult to see the correlations. 

\subsection{Code Size}
\begin{figure}[h!]
    \centering\includegraphics[width=\linewidth]{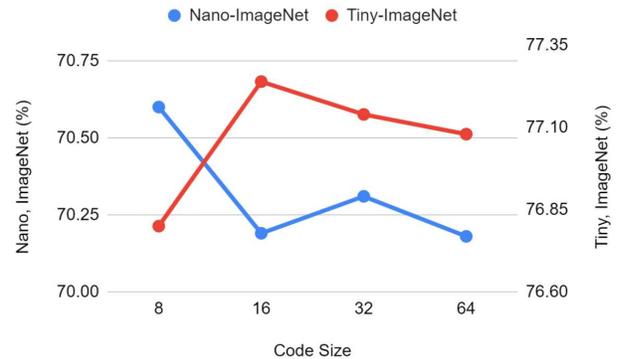}
    \vspace{-7mm}
    \caption{ImageNet classification top-1 accuracy of nano (Nano-ImageNet) and tiny (Tiny-ImageNet) models with respect to the code size from 8 to 64. The nano models correspond to the left axis, and the tiny models correspond to the right axis. Best viewed in color.}
    \vspace{-4mm}
    \label{fig_code_dimension_imagenet}
\end{figure}

It is important to select an appropriate code size to utilize the code ${C}_q$, ${C}_k$, and ${C}_v$ in the F-SNE model.
We conducted the experiments for the code size, from 8 to 64, to find an appropriate code size that leads to better performance.
As shown in Figure~\ref{fig_code_dimension_imagenet}, the code size of 8 achieved the best accuracy between the nano models and 16 achieved the best accuracy between the tiny models.
The tiny models have an embedding dimension of about 1.5 times the embedding dimension of the nano models, as shown in Table~\ref{tab_dimension}.
It can be interpreted that it requires a larger code size when the embedding dimension of the model gets larger.

\section{Limitations}
The limitations of our work are considered in two aspects: performance and constraint. In terms of performance, our approaches are degraded in the Stanford Cars dataset with the XCiT-T12 model, as shown in Table \ref{tab_transfer}. For the constraint, we started with the constraints of the small models. However, our approach can also be applied to other large models, and we believe that small modifications can make our structures easily applicable to large models.

\section{Conclusion}
In this paper, we proposed three new types of $Q, K$, and $V$ vector embedding structures for ViT. The first embedding structure utilizes two non-linear layers to embed the input token into separate non-linear spaces of $Q, K$, and $V$. The second structure shares a single layer between the two layers. Based on experiments, we observed that sharing a single layer is effective for ImageNet classification.
The third structure shares two layers with the $Q, K$, and $V$ codes. The codes are trained via a back-propagation algorithm to minimize the loss of ViT. The structure is helpful for improving the classification rates in several downstream tasks, such as CIFAR-100 and STL-10.
We could easily improve the XCiT model, which is a representative ViT model, using the proposed structures. In particular, our structures performed well with a small number of parameters (approximately 3M or 6M), but we could not prove the effectiveness of the structures in large models. Some modifications may be required to apply the proposed structures to large models; however, we believe that our research is valuable because it can be the starting point for future studies in this direction.

{\small
\bibliographystyle{ieee_fullname}
\bibliography{references}
}

\end{document}